\documentclass[runningheads]{llncs}
\usepackage{graphicx}
\usepackage{comment}
\usepackage{amsmath,amssymb} 
\usepackage{color}
\usepackage{float}
\usepackage{xspace}
\usepackage{diagbox}
\usepackage{multirow}

\usepackage[pagebackref=true,breaklinks=true,letterpaper=true,colorlinks,bookmarks=false]{hyperref}

\usepackage{caption}
\usepackage{subcaption}

\def \OURS {\textit{Neural Voice Puppetry}}

\usepackage[absolute,overlay]{textpos}

\begin{document}

\iftrue
\definecolor{mncolor}{RGB}{255,50,00}
\newcommand\MATTHIAS[1] {\textbf{\textcolor{mncolor}{MN: #1}}}
\definecolor{jtcolor}{RGB}{0,0,255}
\newcommand\JT[1] {\emph{\textcolor{jtcolor}{JT: #1}}}
\definecolor{mzcolor}{RGB}{10,120,10}
\newcommand\MZ[1] {\emph{\textcolor{mzcolor}{MZ: #1}}}
\definecolor{mecolor}{RGB}{250,150,10}
\newcommand\ME[1] {\emph{\textcolor{mecolor}{ME: #1}}}
\definecolor{atcolor}{RGB}{10,120,10}
\newcommand\AT[1] {\emph{\textcolor{atcolor}{AT: #1}}}
\definecolor{todocolor}{RGB}{255,0,00}
\newcommand\TODO[1] {\emph{\textcolor{todocolor}{TODO: #1}}}
\else 
\newcommand\MATTHIAS[1] {}
\newcommand\JT[1] {}
\newcommand\TODO[1] {}
\fi

\definecolor{revcolor}{RGB}{255,50,0}
\definecolor{rev2color}{RGB}{0,50,255}
\newcommand\rev[1] {\emph{\textcolor{revcolor}{#1}}}
\newcommand\REV[1] {\emph{\textcolor{rev2color}{#1}}}

\pagestyle{headings}
\mainmatter
\def\ECCVSubNumber{2619}  

\title{Neural Voice Puppetry: \\Audio-driven Facial Reenactment}

\begin{textblock*}{16cm}(2.4cm,25.0cm) 
   \noindent
   This is a preprint of the accepted version of the following ECCV2020 article: "Neural Voice Puppetry: Audio-driven Facial Reenactment".
\end{textblock*}
\newcommand*{\ShowNotes}{}

\titlerunning{Neural Voice Puppetry}
%
\author{Justus Thies\inst{1} \and
Mohamed Elgharib\inst{2} \and
Ayush Tewari\inst{2} \and
Christian Theobalt\inst{2} \and
Matthias Nie{\ss}ner\inst{1}}
\authorrunning{J. Thies et al.}
%
\institute{Technical University of Munich \and
Max Planck Institute for Informatics, Saarland Informatics Campus}

\maketitle

\begin{figure*}
    \centering
    \includegraphics[width=\linewidth]{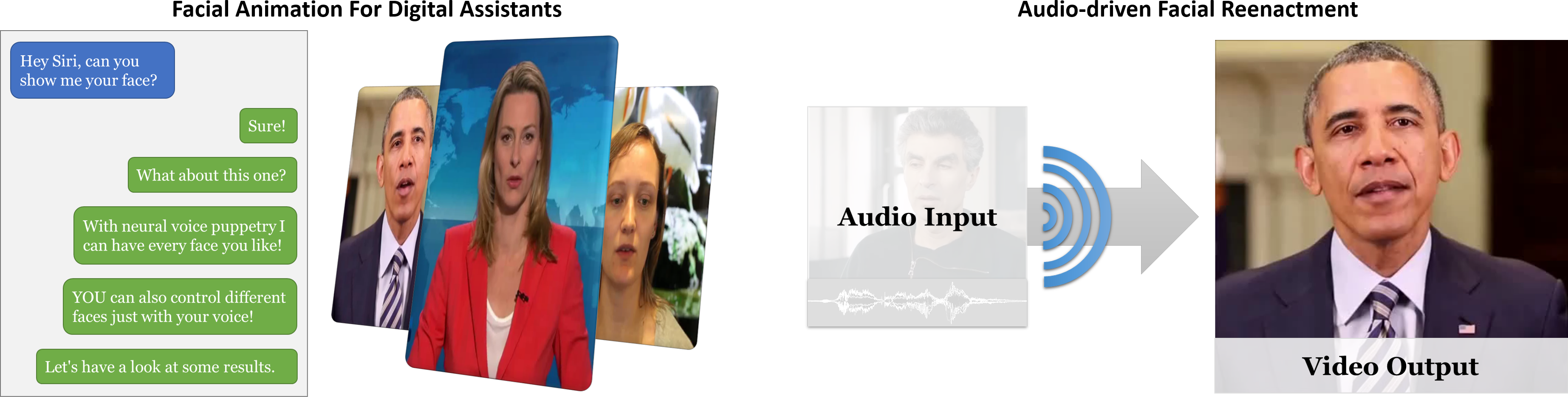}
    \caption{\OURS\xspace enables applications like facial animation for digital assistants or audio-driven facial reenactment.}
    \label{fig:teaser}
\end{figure*}

\begin{abstract}
We present \OURS, a novel approach for audio-driven facial video synthesis\footnote{Video, Code and Demo:\\ \url{https://justusthies.github.io/posts/neural-voice-puppetry/}}.
Given an audio sequence of a source person or digital assistant, we generate a photo-realistic output video of a target person that is in sync with the audio of the source input. 
This audio-driven facial reenactment is driven by a deep neural network that employs a latent 3D face model space.
Through the underlying 3D representation, the model inherently learns temporal stability while we leverage neural rendering to generate photo-realistic output frames.
Our approach generalizes across different people, allowing us to synthesize videos of a target actor with the voice of any unknown source actor or even synthetic voices that can be generated utilizing standard text-to-speech approaches.
\OURS{} has a variety of use-cases, including audio-driven video avatars, video dubbing, and text-driven video synthesis of a talking head.
We demonstrate the capabilities of our method in a series of audio- and text-based puppetry examples, including comparisons to state-of-the-art techniques and a user study.
\end{abstract}

\section{Introduction}
\label{sec:intro}

In the recent years, speech-based interaction with computers made significant progress.
Digital voice assistants are now ubiquitous due to their integration into many commodity devices such as smartphone, tvs, cars, etc.; even companies use more and more machine learning techniques to drive service bots that interact with their customers.
These virtual agents aim for a user-friendly man-machine interface while keeping maintenance costs low.
However, a significant challenge is to appeal to humans by delivering information through a medium that is most comfortable to them.
While speech-based interaction is already very successful, such as shown in virtual assistants like Siri, Alexa, Google, etc., the visual counterpart is largely missing. 
This comes to no surprise given that a user would also like to associate the visuals of a face with the generated audio, similar to the ideas behind video conferencing.
In fact, the level of engagement for audio-visual interactions is higher than for purely audio ones~\cite{Sellen1997,Tarasuik2013}.

The aim of this work is to provide the missing visual channel by introducing \OURS{}, a photo-realistic facial animation method that can be used in the scenario of a visual digital assistant.
To this end, we build on the recent advances in text-to-speech synthesis literature \cite{Jia2018,WaveNet}, which is able to provide a synthetic audio stream from a text that can be generated by a digital agent.
As visual basis, we leverage a short target video of a real person.
The key component of our method is to estimate lip motions that fit the input audio and to render the appearance of the target person in a convincing way.
This mapping from audio to visual output is trained using the ground truth information that we can gather from a target video (aligned real audio and image data).
We designed \OURS{} to be an easy to use audio-to-video translation tool which does not require vast amount of video footage of a single target video or any manual user input.
In our experiments, the target videos are comparably short (2-3 min), thus, allowing us to work on a large amount of video footage that can be downloaded from the Internet. 
To enable this easy applicability to new videos, we generalize specific parts of our pipeline.
Specifically, we compute a latent expression space that is generalized among multiple persons (in our experiments $116$).
This also ensures the capability of being able to handle different audio inputs.
Besides the generation of a visual appearance of a digital agent, our method can also be used as audio-based facial reenactment.
Facial reenactment is the process of re-animating a target video in a photo-realistic manner with the expressions of a source actor~\cite{thies2016face,Zollhoefer2018FaceSTAR}.
It enables a variety of applications, ranging from consumer-level teleconferencing through photo-realistic virtual avatars~\cite{DeepAppearanceModels,thies2018facevr,thies2018headon} to movie production applications such as video dubbing~\cite{Garrido2015,Kim19NeuralDubbing}.
Recently, several authors started to exploit the audio signal for facial reenactment \cite{Chung17b,Suwajanakorn2017,Vougioukas2018EndtoEndSF}.
This has the potential of avoiding failures of visual-based approaches, when the visual signal is not reliable, e.g., due to occluded face, noise, distorted views and so on.
Many of these approaches, however, lack video-realism \cite{Chung17b,Vougioukas2018EndtoEndSF}, since they work in a normalized space of facial imagery (cropped, frontal faces), to be agnostic to head movements.
An exception is the work of Suwajanakorn et al. \cite{Suwajanakorn2017}, where they have shown photo-realistic videos of President Obama that can be synthesized just from the audio signal. 
This approach, however, requires very large quantities of data for training (17 hours of President Obama weekly speeches) and, thus, limits its application and generalization to other identities.
In contrast, our method only needs 2-3 min of a target video to learn the person-specific talking style and appearance.
Our underlying latent 3D model space inherently learns 3D consistency and temporal stability that allows us to generate natural, full frame imagery.
Especially, it enables the disentanglement of rigid head motions from facial expressions.
\noindent
To enable photo-realistic renderings of digital assistants as well as audio-driven facial reenactment, we have the following contributions:
\begin{itemize}
    \item A temporal network architecture called \textit{Audio2ExpressionNet} is proposed to map an audio stream to a 3D blendshape basis that can represent person-specific talking styles.
    Exploiting features from a pre-trained speech-to-text network, we generalize the \textit{Audio2ExpressionNet} on a dataset of news-speaker.

    \item Based on a short target video sequence (2-3 min), we extract a representation of \textit{person-specific talking styles}, since our goal is to preserve the talking style of the target video during reenactment.

    \item A novel \textit{light-weight neural rendering network} using neural textures is presented that allows us to generate photo-realistic video content reproducing the person-specific appearance. It surpasses the quality and speed of state-of-the-art neural rendering methods~\cite{Fried2019,thies2019}.
    
\end{itemize}

\section{Related Work}
\label{sec:related}
\OURS{} is a facial reenactment approach based only on audio input.
In the literature, there are many video-based facial reenactment systems that enable dubbing and other general facial expression manipulation.
Our focus in this related work section lies on audio-based methods.
These methods can be organized in facial animation and facial reenactment.
Facial animation concentrates on the prediction of expressions that can be applied to a predefined avatar.
In contrast, audio-driven facial reenactment aims to generate photo-realistic videos of an existing person including all idiosyncrasies. 

\noindent
\textbf{Video-Driven Facial Reenactment:}
The state-of-the-art report of Zollh\"ofer et al.~\cite{Zollhoefer2018FaceSTAR} discusses several works for video-driven facial reenactment.
Most methods, rely on a reconstruction of a source and target face using a parametric face model.
The target face is reenacted by replacing its expression parameters with that of the source face.
Thies et al. \cite{thies2016face} uses a static skin texture and a data-driven approach to synthesize the mouth interior.
In Deep Video Portraits~\cite{kim2018DeepVideo}, a generative adversarial network is used to produce photo-realistic skin texture that can handle skin deformations conditioned on synthetic renderings.
Recently, Thies et al. \cite{thies2019} proposed the usage of neural textures in conjunction with a deferred neural renderer.
%
Results show that neural textures can be used to generate high quality facial reenactments.
For instance, it produces high fidelity mouth interiors with less artifacts.
Kim et al. \cite{Kim19NeuralDubbing} analyzed the notion of style for facial expressions and showed its importance for dubbing.
%
In contrast to Kim et al. \cite{Kim19NeuralDubbing}, we directly estimate the expressions in the target talking-style domain, thus, we don't need to apply any transfer or adaption method.

\noindent
\textbf{Audio-Driven Facial Animation:}
These methods do not focus on photo-realistic results, but on the prediction of facial motions~\cite{VOCA2019,Karras2017,PhamCP17,Taylor2017,Tzirakis19}.
Karras et al. \cite{Karras2017} drives a 3D facial animation using an LSTM that maps input waveforms to the 3D vertex coordinates of a face mesh, also considering the emotional state of the person. In contrast to our method, it needs high quality 3D reconstructions for supervised training and does not render photo-realistic output.
Taylor et al. \cite{Taylor2017} use a neural network to map phonemes into the parameters of a reference face model. It is trained on data collected for only one person speaking for 8 hours. They show animations of different synthetic avatars using deformation retargeting.
VOCA~\cite{VOCA2019} is an end-to-end deep neural network for speech-to-animation translation trained on multiple subjects. Similar to our approach, a low-dimensional audio embedding based on features of the DeepSpeech network \cite{deepspeech} is used. From this embedding, VOCA regresses 3D vertices on a FLAME face model \cite{flame} conditioned on a subject label. It requires high quality 4D scans recorded in a studio setup. Our approach works on 'in the wild' videos, with a focus on temporally coherent predictions and photo-realistic renderings.
\noindent
\textbf{Audio-Driven Facial Reenactment:}
Audio-driven facial reenactment has the goal to generate photo-realistic videos that are in sync with the input audio stream.
There is a number of techniques for audio-driven facial reenactment \cite{Bregler1997,chen2019hierarchical,Chung17b,Ezzat2002,Vougioukas2018EndtoEndSF,Vougioukas2019RealisticSF} but only a few generate photo-realistic, natural, full frame images~\cite{Suwajanakorn2017}.
Suwajanakorn et al.~\cite{Suwajanakorn2017} uses an audio stream from President Barack Obama to synthesize a high quality video of him speaking.
A Recurrent Neural Network is trained on many hours of his speech to learn the mouth shape from the audio.
The mouth is then composited with proper 3D matching to reanimate an original video in photo-realistic manner.
Because of the huge amount of used training data (17h), it is not applicable to other target actors.
In contrast, our approach only needs a 2-3 min long video of a target sequence.
Chung et al. \cite{Chung17b} present a technique that animates the mouth of a still, normalized image to follow an audio speech.
First, the image and audio is projected into a latent space through a deep encoder.
A decoder then utilizes the joint embedding of the face and audio to synthesize the talking head.
The technique is trained on tens of hours of data in an unsupervised manner.
Another 2D image-based method has been presented by Vougioukas et al. \cite{Vougioukas2018EndtoEndSF}.
They use a temporal GAN to produce a video of a talking face given a still image and an audio signal as input.
The generator feeds the still image and the audio to an encoder-decoder architecture with a RNN to better capture temporal relations.
It uses discriminators that work on per-frame and on a sequence level to improve temporal coherence.
As conditioning, it also takes the audio signal as input to enforce the synthesized mouth to be in sync with the audio.
In \cite{Vougioukas2019RealisticSF} a dedicated mouth-audio syn discriminator is used to improve the results. 
In contrast to our method, the 2D image-based methods are restricted to a normalized image space of cropped and frontalized images.
They are not applicable to generate full frame images with 3D consistent motions.
\noindent
\textbf{Text-Based Video Editing:}
Fried et al. \cite{Fried2019} presented a technique for text-based editing of videos.
Their approach allows overwriting existing video segments with new texts in a seamless manner.
A face model \cite{garrido2016reconstruction} is registered to the examined video and a viseme search finds video segments with similar mouth movements to the edited text.
The corresponding face parameters of the matching video segment are blended with the original sequence parameters based on a heuristic, followed by a deep renderer to synthesize photo-realistic results.
The method is person-specific and requires a one hour long training sequence of the target actor and, thus, is not applicable to short videos from the Internet.
The viseme search is slow ($\sim5$min for three words) and does not allow for interactive results.

\section{Overview}
\label{sec:overview}
\begin{figure*}[t!]
    \centering
    \includegraphics[width=\linewidth]{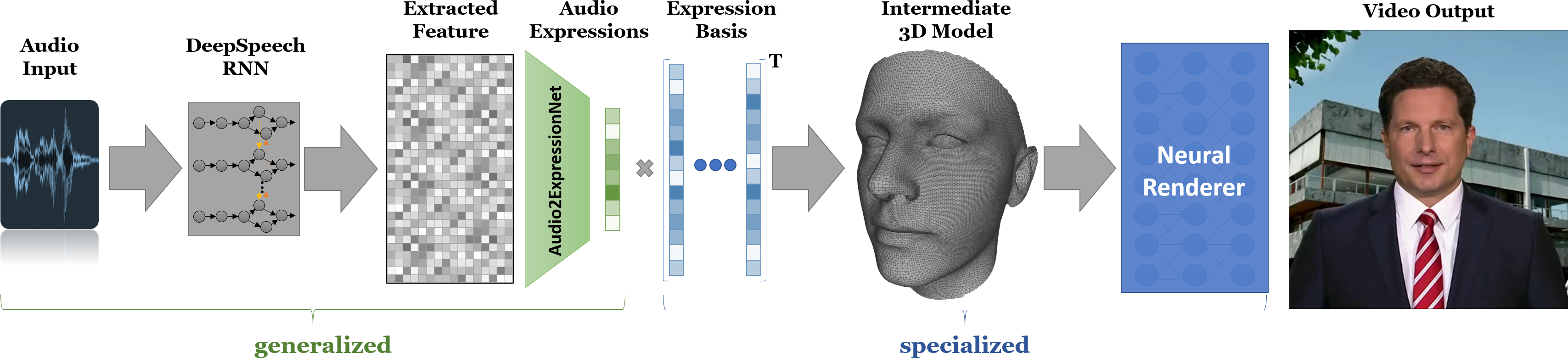}
    \caption{Pipeline of \OURS. Given an audio sequence we use the DeepSpeech RNN to predict a window of character logits that are fed into a small network.
    This generalized network predicts coefficients that drive a person-specific expression blendshape basis.
    We render the target face model with the new expressions using a novel light-weight neural rendering network.
    }
    \label{fig:overview}
\end{figure*}

\OURS\xspace consists of two main components (see Fig.~\ref{fig:overview}): a generalized and a specialized part.
A generalized network predicts a latent expression vector, thus, spanning an \textit{audio-expression space}.
This \textit{audio-expression space} is shared among all persons and allows for reenactment, i.e., transferring the predicted motions from one person to another.
To ensure generalizability w.r.t. the input audio, we use features extracted by a pretrained speech-to-text network~\cite{deepspeech} as input to estimate the audio-expressions.
The audio-expressions are interpreted as blendshape coefficients of a 3D face model rig.
This face model rig is person-specific and is optimized in the second part of our pipeline.
This specialized stage captures the idiosyncrasies of a target person including the facial motion and appearance.
It is trained on a short video sequence of $2-3$ minutes (in comparison to hours that are required by state-of-the-art methods).
The 3D facial motions are represented as delta-blendshapes which we constrain to be in the subspace of a generic face template~\cite{Blanz1999,thies2016face}.
A neural texture in conjunction with a novel neural rendering network is used to store and to rerender the appearance of the face of an individual person.
%

\begin{figure}
    \centering
    \includegraphics[trim={0 0 0 8.75cm},clip,width=\linewidth]{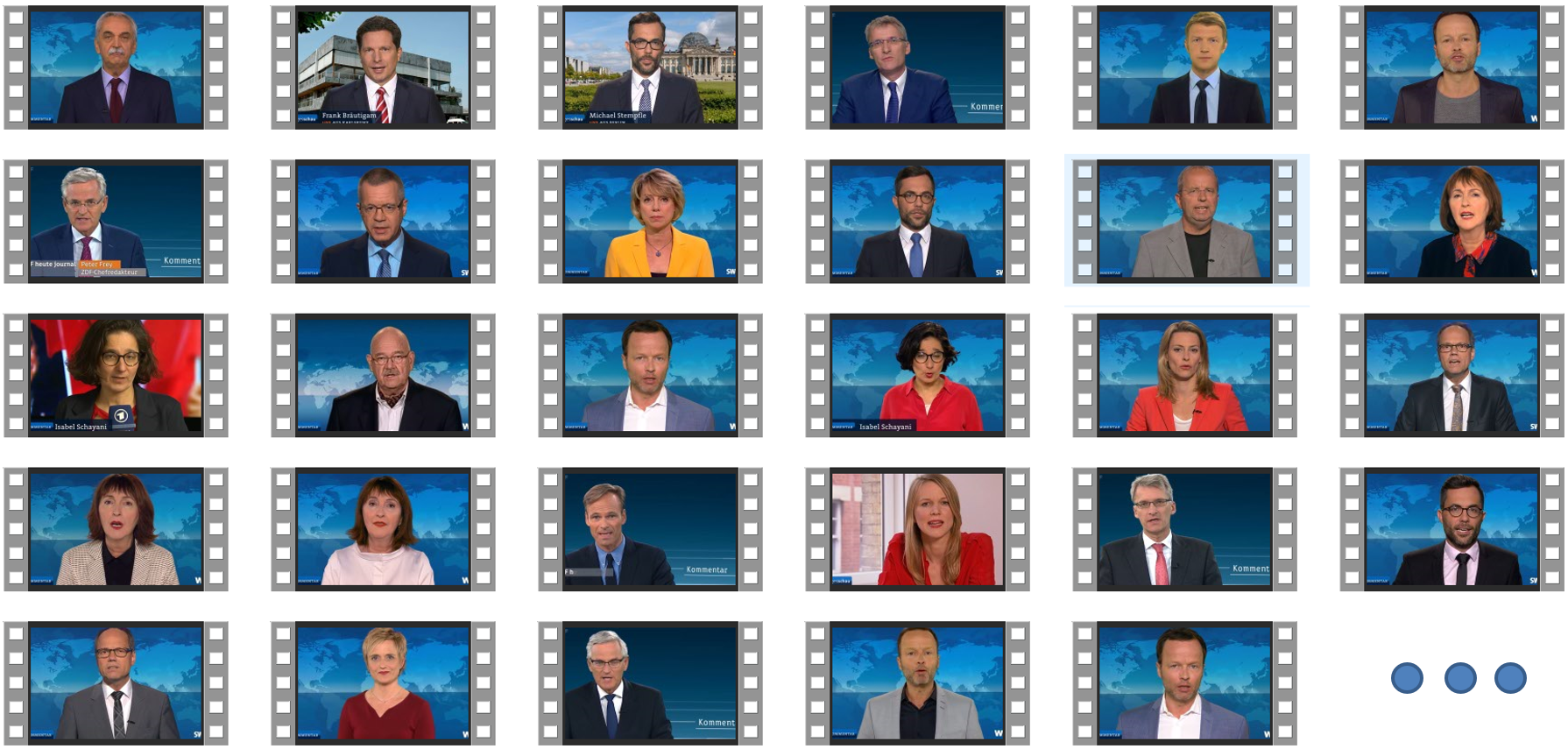}
    \caption{Samples of the training corpus used to optimize the \textit{Audio2ExpressionNet}.}
    \label{fig:training_corpus}
\end{figure}

\section{Data}
\label{sec:data}

In contrast to previous model-based methods, \OURS\xspace is based on \textit{'in-the-wild' videos that can be download from the internet}.
%
The videos have to be synced with the audio stream, such that we can extract ground truth pairs of audio features and image content.
In our experiments the videos have a resolution of $512 \times 512$ with $25fps$.

\noindent
\textbf{Training Corpus for the \textit{Audio2ExpressionNet}:}
Fig.~\ref{fig:training_corpus} shows an overview of our video training corpus that is used for the training of the small network that predicts the 'audio expressions' from the input audio features (see Sec.~\ref{sec:audio2expression}).
The dataset consists of $116$ videos with an average length of $1.7min$ (in total $302750$ frames).
We selected the training corpus, such that the persons are in a neutral mood (commentators of the German public TV).

\noindent
\textbf{Target Sequences:}
For a target sequence, we extract the person-specific talking style in the sequence.
I.e., we compute a mapping from the generalized audio-expression space to the actual facial movements of the target actor (see Sec.~\ref{sec:training}).
The sequences are $2-3min$ long and, thus, easy to obtain from the Internet.
%
%

\subsection{Preprocessing:}
In an automatic preprocessing step, we extract face tracking information as well as audio features needed for training.

\noindent
\textbf{3D Face Tracking: }
Our method is using a statistical face model and delta-blendshapes~\cite{Blanz1999,thies2016face} to represent a 3D latent space for modelling facial animation.
The 3D face model space reduces the face space to only a few hundred parameters ($100$ for shape, $100$ for albedo and $76$ for expressions) and stays fixed in this work.
Using the dense face tracking method of Thies et al.~\cite{thies2016face}, we estimate the model parameters for every frame of a sequence.
%
%
During tracking, we extract the per-frame expression parameters that are used to train the audio to expression network.
To train our neural renderer, we also store the rasterized texture coordinates of the reconstructed face mesh.

\noindent
\textbf{Audio-feature Extraction:}
The video contains a synced audio stream.
We use the recurrent feature extractor of the pre-trained speech-to-text model DeepSpeech~\cite{deepspeech} (v0.1.0).
Similar to Voca~\cite{VOCA2019}, we extract a window of character logits per video frame.
Each window consists of $16$ time intervals \`a $20$ms, resulting in an audio feature of $16\times29$.
The DeepSpeech model is generalized among thousands of different voices, trained on Mozilla's CommonVoice dataset.

\section{Method}
\label{sec:main}
To enable photo-realistic facial reenactment based on audio signals, we employ a 3D face model as intermediate representation of facial motion.
A key component of our pipeline is the audio-based expression estimation.
Since every person has his own talking style and, thus, different expressions, we establish person-specific expression spaces that can be computed for every target sequence.
To ensure generalization among multiple persons, we created a latent \textit{audio-expression space} that is shared by all persons.
From this audio-expression space, one can map to the person specific expression space, enabling reenactment.
Given the estimated expression and the extracted audio features, we apply a novel light-weight neural rendering technique that generates the final output image.
\subsection{Audio2ExpressionNet}
\label{sec:audio2expression}

Our method is designed to generate temporally smooth predictions of facial motions.
To this end, we employ a deep neural network with two stages.
First, we predict per-frame facial expression predictions.
These expressions are potentially noisy, thus, we use an expression aware temporal filtering network.
Given the noisy per-frame predictions as input the neural network predicts filter weights to compute smooth audio-expressions for a single frame.
The per-frame as well as the filtering network can be trained jointly and outputs audio-expression coefficients.
This audio-expression space is shared among multiple persons and is interpreted as blendshape coefficients.
Per person, we compute a blendshape basis which is in the subspace of our generic face model~\cite{thies2016face}.
The networks are trained with a loss that works on a vertex level of this face model.

\subsubsection{Per-frame Audio-Expression Estimation Network:}
Since our goal is a generalized audio-based expression estimation, we rely on generalized audio features.
We use the RNN-part of the speech to text approach DeepSpeech~\cite{deepspeech} to extract these features.
These features represent the logits of the DeepSpeech alphabet for $20ms$ audio signal.
For each video frame, we extract a time window of $16$ features around the frame that consist of $29$ logits (length of the DeepSpeech alphabet is $29$).
This, $16 \times 29$ tensor is input to our per-frame estimation network (see Fig.~\ref{fig:expression_filtering}).
To map from this feature space to the per-frame audio-expression space, we apply $4$ convolutional layer and $3$ fully connected layer.
Specifically, we apply 1D convolutions with kernel dimensions $(3)$ and stride $(2)$, filtering in the time dimension. The convolutional layers have a bias and are followed by a leaky ReLU (slope $0.02$).
The feature dimensions are reduced successively from $(16\times29)$,$(8\times32)$,$(4\times32)$,$(2\times64)$ to $(1\times64)$.
This reduced feature is input to the fully connected layers that have a bias and are also followed by a leaky ReLU ($0.02$), except the last layer.
The fully connected layers map the $64$ features from the convolutional network to $128$, then to $64$ and, finally, to the audio-expression space of dimension $32$, where a TanH activation is applied.

\subsubsection{Temporally Stable Audio-Expression Estimation:}

\begin{figure*}[t]
    \centering
    \begin{subfigure}[t]{0.6\linewidth}
        \centering
        \includegraphics[height=3.2cm]{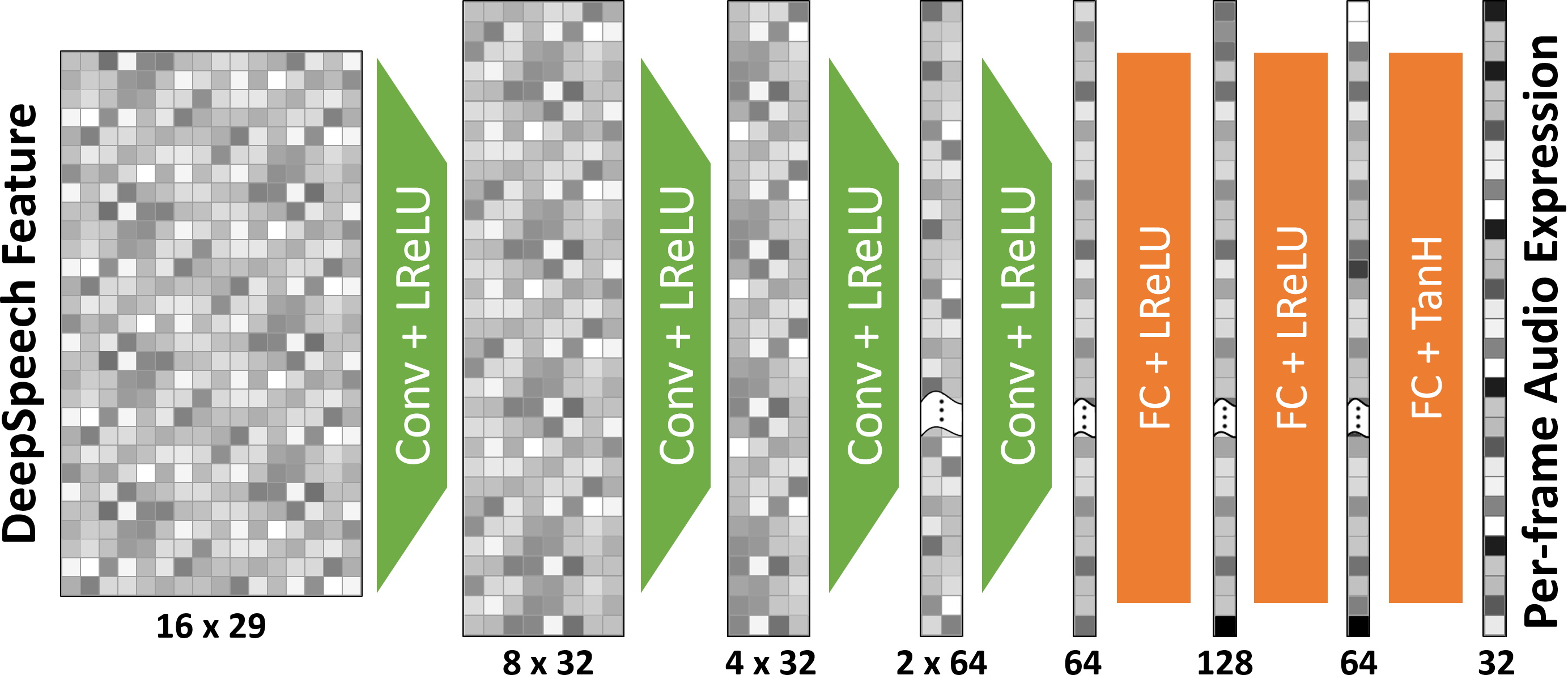}
        \caption{}
    \end{subfigure}%
    ~~~~~
    \begin{subfigure}[t]{0.35\linewidth}
        \centering
        \includegraphics[height=3.2cm]{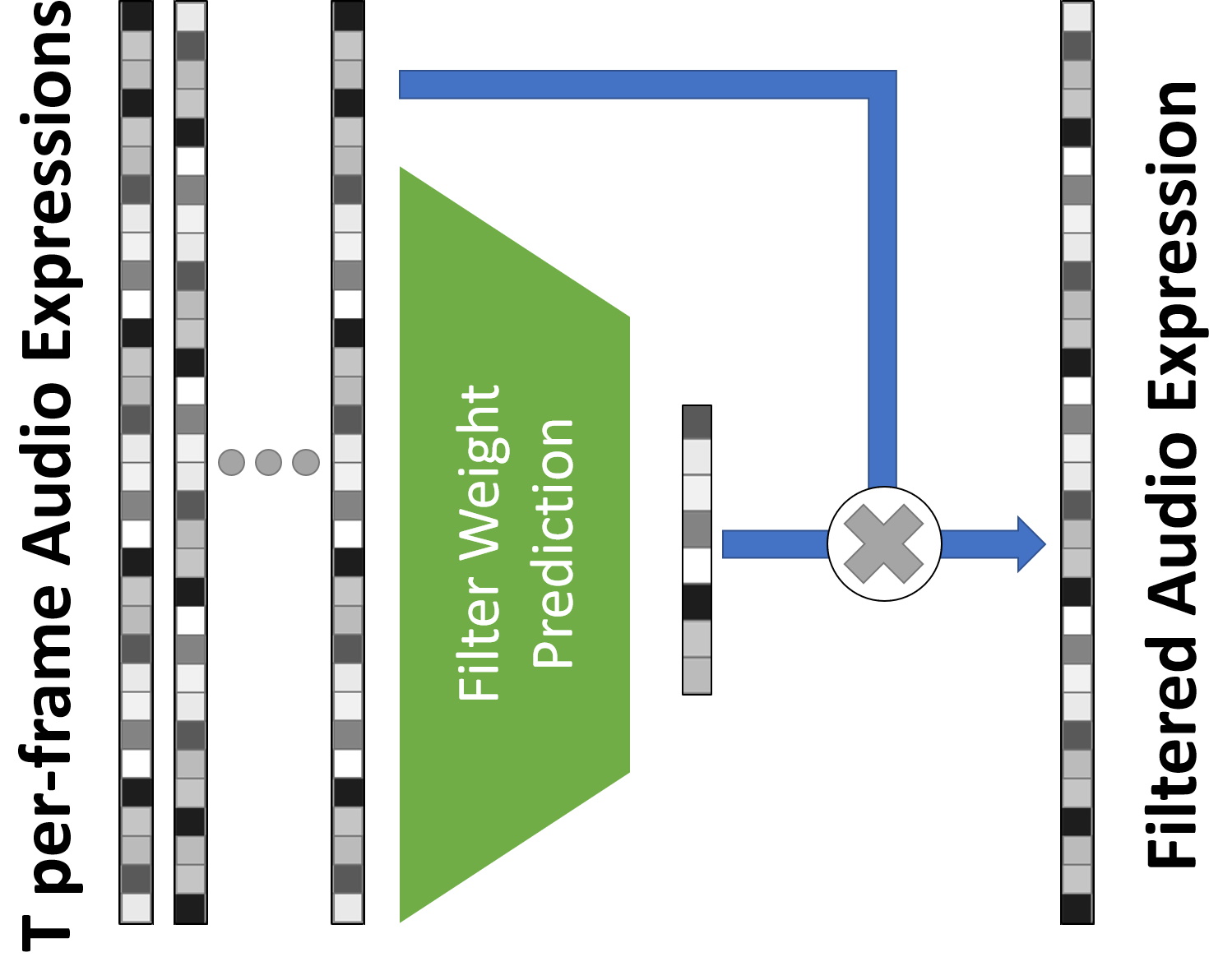}
        \caption{}
    \end{subfigure}

    \caption{Audio2ExpressionNet: (a) Per-frame audio-expression estimation network that gets DeepSpeech features as input, (b) to get smooth audio-expressions, we employ a content-aware filtering along the time dimension.}
    \label{fig:expression_filtering}
\end{figure*}

To generate temporally stable audio-expression predictions, we jointly learn a filtering network that gets $T$ per-frame estimates as input (see Fig.~\ref{fig:expression_filtering} (b)).
Specifically, we estimate the audio-expressions for frame $t$ using a linear combination of the per-frame predictions of the timesteps $t-T/2$ to $t+T/2$.
The weights for the linear combination are computed using a neural network that gets the audio-expressions as input (which results in an expression-aware filtering).
The filter weight prediction network consists of five 1D convolutions followed by a linear layer with softmax activation (see supplemental material for detailed description).
This content aware temporal filtering is also inspired by the self-attention mechanism~\cite{Han18}.

\subsubsection{Person-specific Expressions:}
To retrieve the 3D model from this audio-expression space, we learn a person-specific audio-expression blendshape basis which we constrain by the generic blendshape basis of our statistical face model.
I.e., the audio-expression blendshapes of a person are a linear combination of the generic blendshapes.
This linear relation, results in a linear mapping from the audio-expression space which is output of the generalized network to the generic blendshape basis.
This linear mapping is person specific, resulting in $N$ matrices with dimension $76 \times 32$ during training ($N$ being the number of training sequences and $76$ being the number of generic blendshapes).

\subsubsection{Loss:}
The network and the mapping matrices are learned end-to-end using the visually tracked training corpus and a vertex-based loss function, with a higher weight (10x) on the mouth region of the face model.
Specifically, we compute a vertex-to-vertex distance from the audio-based predicted and the visually tracked face model in terms of a root mean squared (RMS) distance:
\begin{equation*}
    L_{expr} = RMS(v_t -  v_t^{*}) + \lambda \cdot L_{temp}
\end{equation*}
with $v_t$, the vertices based on the filtered expression estimation of frame $t$ and $v_t^{*}$ being the visual tracked face vertices.
In addition to the absolute loss between predictions and the visual tracked face geometry, we use a temporal loss that considers the vertex displacements of consecutive frames:
\begin{equation*}
\begin{split}
\footnotesize
    L_{temp} &= RMS((v_t-v_{t-1})      -  (v_t^{*}-v_{t-1}^{*}))
             + RMS( (v_{t+1}-v_{t})   -  (v_{t+1}^{*}-v_{t}^{*})) \\
             &+ RMS( (v_{t+1}-v_{t-1}) -  (v_{t+1}^{*}-v_{t-1}^{*}))
\end{split}
\end{equation*}
These forward, backward and central differences are weighted with $\lambda$ (in our experiments $\lambda=20$).
The losses are measured in millimeters.
\subsection{Neural Face Rendering}

Based on the recent advances in neural rendering, we employ a novel light-weight neural rendering technique that is based on neural textures to store the appearance of a face.
~Our rendering pipeline synthesizes the lower face in the target video based on the audio-driven expression estimations.
Specifically, we use two networks (see supplemental material for an overview figure).
One network that focuses on the face interior, and another network that embeds this rendering into the original image.
The estimated 3D face model is rendered using the rigid pose observed from the original target image using a neural texture~\cite{thies2019}. The neural texture has a resolution of $256\times256\times16$.
The network for the face interior translates these rendered feature descriptors to RGB colors.
The network is using a similar structure as a classical U-Net with 5 layers.
But instead of using strided convolutions that result in a downsampling in each layer, we are using dilated convolutions with increasing dilation factor and a stride of one. Instead of transposed convolutions we are using standard convolutions.
All convolutions have kernel size $3\times3$.
Note, dilated instead of strided convolutions do not increase the number of learnable parameters, but it increases the memory load during training and testing.
Dilated convolutions reduce visual artifacts and result in smoother results (also temporally, see video).
The second network that blends the face interior with the 'background image' has the same structure.
To remove potential movements of the chin in the background image, we erode the background image around the rendered face.
The second network inpaints these missing regions.

\subsubsection{Loss:}

We use a per-frame loss function that is based on an $\ell_1$ loss to measure absolute errors and a VGG style loss~\cite{Johnson2016}.
\begin{equation*}
    L_{rendering} = \ell_1(I, I^{*}) + \ell_1(\hat{I}, I^{*}) + VGG(I, I^{*})
\end{equation*}
with $I$ being the final synthetic image, $I^*$ the ground truth image and $\hat{I}$ the intermediate result of the first network that focuses on the face interior (loss is masked to this region).

\subsection{Training}
\label{sec:training}

Our training procedure has two stages -- the generalization and the specialization phase.
In the first phase, we train the \textit{Audio2ExpressionNet} among all sequences from our dataset (see Sec.~\ref{sec:data}) in a supervised fashion.
Given the visual face tracking information, we know the 3D face model of a specific person for every frame.
In the training process, we reproduce these 3D reconstructions based on the audio input by optimizing the network parameters and the person-specific mappings from the audio-expression space to the 3D space.
In the second phase, the rendering network for a specific target sequence is trained.
Given the ground truth images and the visual tracking information, we train the neural renderer end-to-end including the neural texture.

\subsubsection{New Target Video:}
Since the audio-based expression estimation network is generalized among multiple persons, we can apply it to unseen actors.
The person specific mapping between the predicted audio-expression space coefficients and the expression space of the new person can be obtained by solving a linear system of equations.
Specifically, we extract the audio-expression for all training images and compute the linear mapping to the expressions that are visually estimated.
In addition to this step, the person-specific rendering network for the new target video is trained from scratch (see supplement for further information).

\subsection{Inference}
At test time, we only require a source audio sequence.
Based on the target actor selection, we use the corresponding person-specific mapping.
The mapping from the audio features to the person specific expression space takes less than $2ms$ on an Nvidia 1080Ti.
Generation of the 3D model and the rasterization using these predictions takes another $2ms$.
The deferred neural rendering takes $\sim5ms$ which results in a real-time capable pipeline.

\subsubsection{Text-to-Video:}

Our pipeline is trained on real video sequences, where the audio is in sync with the visual content.
Thus, we learned a mapping directly from audio to video that ensures synchronicity.
Instead of going directly from text to video, where such a natural training corpus is not available, we synthesize voice from the text and feed this into our pipeline.
For our experiments we used samples from the DNN-based text-to-speech demo of IBM Watson\footnote{\url{https://text-to-speech-demo.ng.bluemix.net/}}.
Which gives us state-of-the-art synthetic audio streams that are comparable to the synthetic voices of  virtual assistants.

\section{Results}
\label{sec:results}
\OURS\xspace has several important use cases, i.e., audio-driven video avatars, video dubbing and text-driven video synthesis of a talking head, see supplemental video.
In the following sections, we discuss these results including comparisons to state-of-the-art approaches.

\begin{figure}
    \centering
    \includegraphics[width=\linewidth]{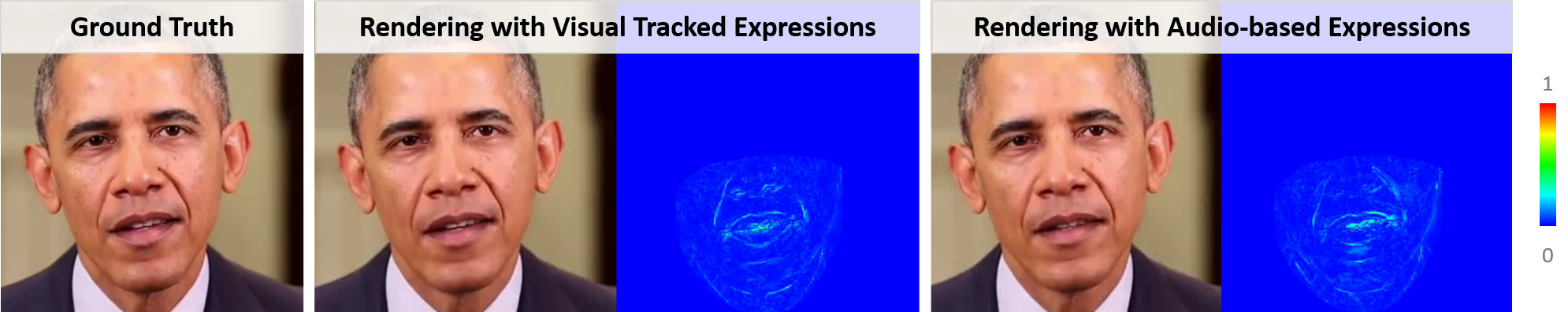}
    \caption{Self-reenactment: Evaluation of our rendering network and the audio-prediction network. Error plot shows the euclidean photo-metric error.}
    \label{fig:self_reenactment}
\end{figure}

\subsection{Ablation studies}

\subsubsection{Self-reenactment:}
We use self-reenactment to evaluate our pipeline (Fig.~\ref{fig:self_reenactment}), since it gives us access to a ground truth video sequence where we can also retrieve visual face tracking.
As a distance measurement, we use an $\ell_2$ distance in color space (colors in [0,1]) and the corresponding PSNR values.
Using this measure, we evaluate the rendering network and the entire reenactment pipeline.
Specifically, we compare the results using visual tracked mouth movements to the results using audio-based predictions (see video).
The mean color difference of the re-rendering on the test sequence of $645$ frame is $0.003$ for the visual and $0.005$ for the audio-based expressions, which corresponds to a PSNR of $41.48$ and $36.65$ respectively.
In addition to the photo-metric measurements, we computed the 2D mouth landmark distances relative to the eye distance using Dlib, resulting in $0.022$ for visual tracking and $0.055$ for the audio-based predictions.
In the supplemental video, we also show a side-by-side comparison of our rendering network using dilated convolutions and our network with strided convolutions (and a kernel size of $4$ to reduce block artifacts in the upsampling).
Both networks are trained with the same number of epochs ($50$).
As can be seen, dilated convolutions lead to visually more pleasing results (smoother in spatial and temporal domain).
As a comparison to the results using dilated convolutions reported above, strided convolutions result in a lower PSNR of $40.12$ with visual tracking and $36.32$ with audio-based predictions.

\noindent
\textbf{Temporal smoothness:}
We also evaluated the benefits of using a temporal-based expression prediction network. Besides temporally smooth predictions shown in the supplemental video, it also improves the prediction accuracy of the mouth shape. The relative 2D mouth landmark error improves from $0.058$ (per frame prediction) to $0.055$ (temporal prediction).

\noindent
\textbf{Generalization / Transferability:}
Our results are covering different target persons which demonstrates the wide applicability of our method,
including the reproduction of different person-specific talking styles and appearances.
As can be seen in the supplemental video, the expression estimation network that is trained on multiple target sequences ($302750$ frames) results in more coherent predictions than the network solely trained on a sequence of Obama ($3145$ frames).
The usage of more target videos increases the training corpus size and the variety of input voices and, thus, leads to more robustness.
In the video, we also show a comparison of the transfer from different source languages to different target videos that are originally also in different languages.
In Tab.~\ref{tab:languages}, we show the corresponding quantitative measurements of the achieved lip sync using SyncNet~\cite{syncnet}.
SyncNet is trained on the BBC news program (English), nevertheless, the authors state that it works also good for other languages.
As a reference for the measurements of the different videos, we list the values for the original target videos.
Higher confidence values are better, while a value below $1$ refers to uncorrelated audio video streams.
The original video of Macron has the lowest measured confidence which propagates to the reenactment results.

\begin{table}[t]
\begin{center}
\resizebox{\textwidth}{!}{%
 \begin{tabular}{|c|| c | c | c | c| c| c|| c|} 
     \hline
     \multirow{2}{*}{\diagbox{~~~Target~~~}{~~~~Source~~~}} & Bengali & Chinese & German & Greek & Spanish & English & Reference \\
     & (Male) & (Female) & (Female) & (Male) & (Male) & (Male) & (Original)\\
     
     \hline\hline
     Obama (English) & ($-3$/$5.137$) & ($-3$/$3.234$)  & ($-3$/$5.544$) & ($-4$/$1.952$) & ($-3$/$4.179$) & ($-3$/$4.927$) & ($-3$/$7.865$)\\ 
     Macron (French) & ($-3$/$3.994$) & ($-3$/$2.579$)  & ($-2$/$3.012$) & ($-3$/$1.856$) & ($-3$/$3.752$) & ($-3$/$3.163$) & ($-1$/$3.017$)\\ 
     News-speaker (German) & ($-2$/$5.361$) & ($-2$/$6.505$)  & ($-2$/$5.734$) & ($-2$/$5.752$) & ($-2$/$6.408$) & ($-2$/$6.036$) & ($-1$/$9.190$) \\ 
     Woman (English) & ($-1$/$6.265$) & ($-1$/$4.431$)  & ($-1$/$3.841$) & ($-1$/$4.206$) & ($-1$/$3.684$) & ($-1$/$4.716$) & ($-1$/$6.036$) \\ 
     \hline
 \end{tabular}
 }
 \end{center}
 \caption{Analysis of generated videos with different source/target languages. Based on SyncNet~\cite{syncnet}, we measure the audio-visual sync (offset/confidence). As a reference, we list the sync measurements for the original target video (right). }
 \label{tab:languages}
\end{table}


\subsection{Comparisons to state-of-the-art methods}


In the following, as well as in the supplemental document, we compare to model-based and pure image-based approaches for audio-driven facial reenactment. \\

\begin{figure}[t]
    \centering
    \includegraphics[width=0.85\linewidth]{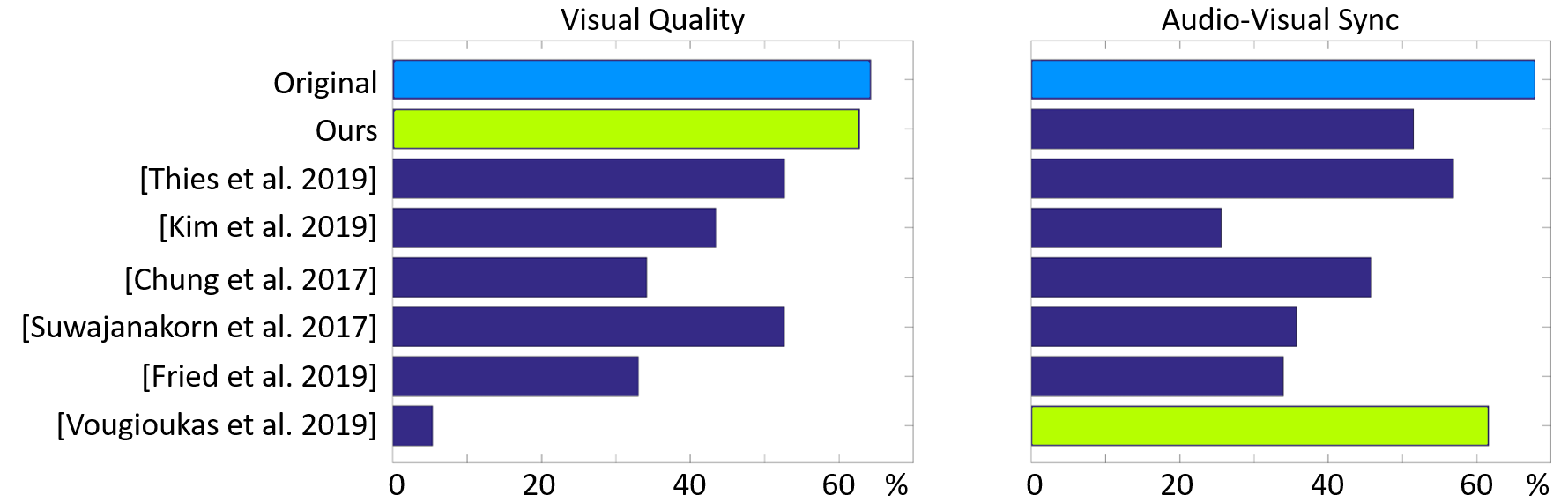}
    \caption{User study: percentage of attendees (in total $56$) that rated the visual and audio-visual quality good or very good.}
    \label{fig:userstudy}
\end{figure}

\noindent
\textbf{Preliminary User Study:}
In a preliminary user study, we evaluated the visual quality and audio-visual sync of the state-of-the-art methods.
The user study is based on videos taken from the supplemental materials of the respective publications (assuming the authors showing the best case scenario).
Note that the videos of the different methods show (potentially) different persons (see supplemental material).
In total, $56$ attendees with a computer science background judged upon synchronicity and visual quality ('very bad', 'bad', 'neither bad nor good', 'good', 'very good') of $24$ videos in randomized order (in total).
In Fig.~\ref{fig:userstudy}, we show the percentage of attendees that rated the specific approach with good or very good.
As can be seen, the 2D image-based approaches achieve a high audio-visual sync (especially, Vougioukas~\cite{Vougioukas2019RealisticSF}), but they lack visual quality and are not able to synthesize natural videos (outside of the normalized space).
Our approach gives the best visual quality and also a high audio-visual sync, similar to state-of-the-art video-based reenactment approaches like Thies et al.~\cite{thies2019}. \\

\begin{figure*}
    \centering
    \begin{subfigure}[t]{0.5\linewidth}
        \centering
        \includegraphics[height=2.35cm]{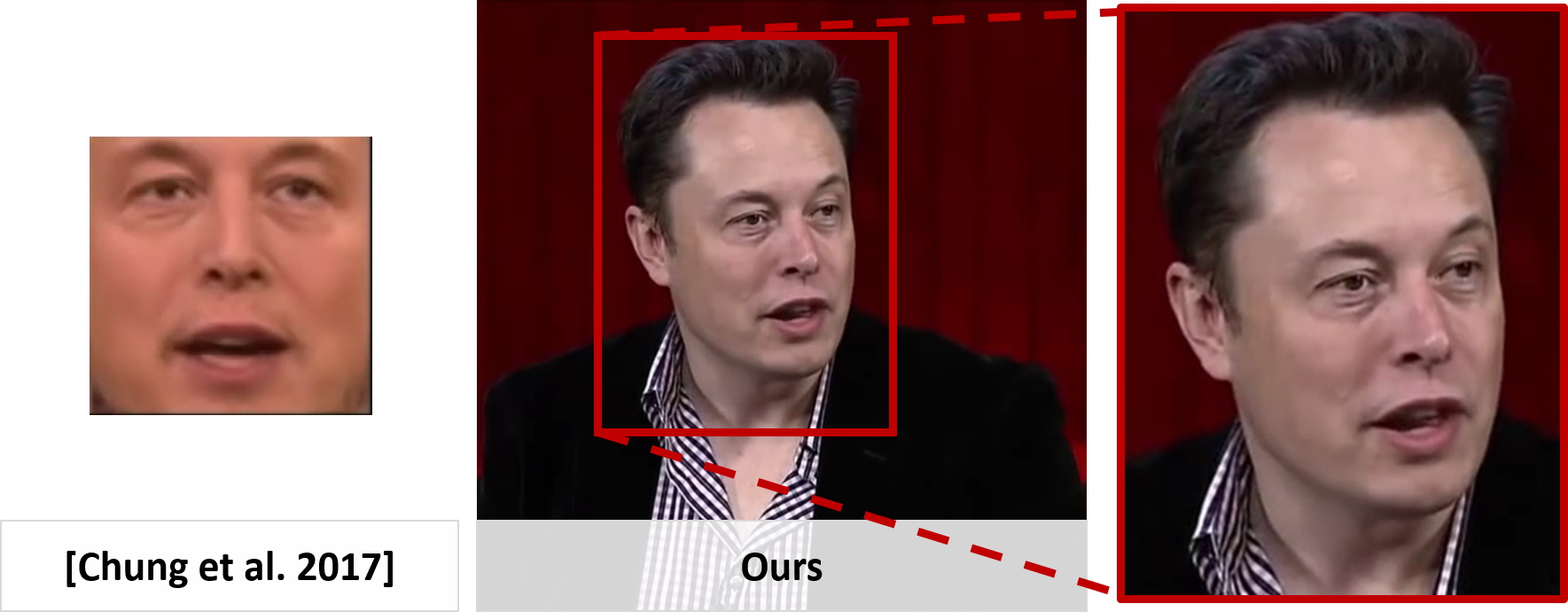}
        \resizebox{!}{0.35cm}{%
         \begin{tabular}{|c| c | c |} 
             \hline
             ~[Chung et al. 2017]~ & Ours & Reference \\
             \hline
             ($-2$/$6.219$) & ~($-2$/$4.538$)~ & ~($-1$/$5.811$)~ \\ 
             \hline
         \end{tabular}
         }
        \caption{}
    \end{subfigure}%
    ~ 
    \begin{subfigure}[t]{0.5\linewidth}
        \centering
        \includegraphics[height=2.35cm]{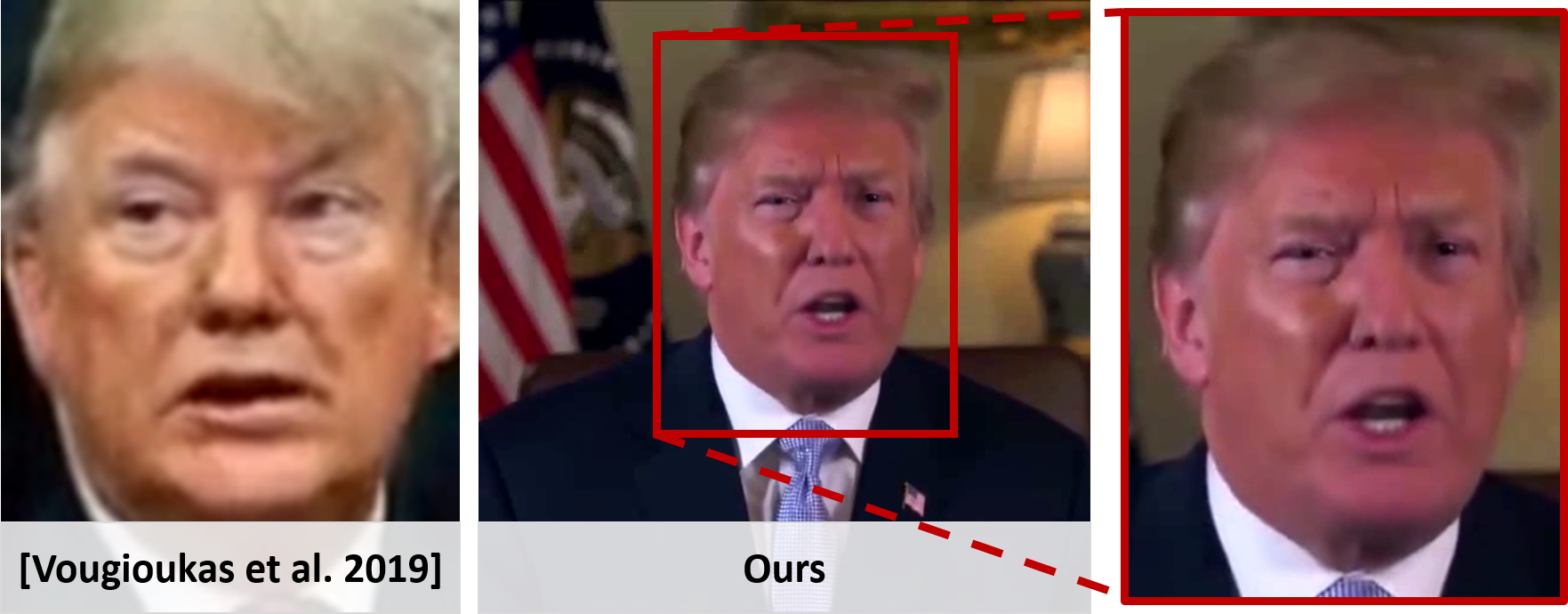}
        \resizebox{!}{0.35cm}{%
         \begin{tabular}{|c| c | c |} 
             \hline
             ~[Vougioukas et al. 2019]~ & Ours & Reference \\
             \hline
             ($-2$/$7.607$)  & ~($-3$/$5.883$)~ & ~($-6$/$4.859$)~ \\ 
             \hline
         \end{tabular}
         }
        \caption{}
    \end{subfigure}
    \caption{Visual quality comparison to the image-based methods (a) 'You said that?'~\cite{Chung17b} and (b) 'Realistic Speech-Driven Facial Animation with GANs'~\cite{Vougioukas2019RealisticSF} (driven by the same input audio stream, respectively), including the synchronicity measurements using SyncNet~\cite{syncnet} (offset/confidence).}
    \label{fig:2d_based_methods}
\end{figure*}

\noindent
\textbf{Image-based Methods:}
Our method aims for high quality output that is embedded in a real video, including the person-specific talking style, exploiting an explicit 3D model representation of the face to ensure 3D consistent movements.
This is fundamentally different from image-based approaches that are operating in a normalized space of facial imagery (cropped, frontal faces) and do not capture person-specific talking styles, but, therefore, can be applied to single input images.
In Fig.~\ref{fig:2d_based_methods} as well as in the video, we show generated images of state-of-the-art image-based methods~\cite{Chung17b,Vougioukas2019RealisticSF}.
It illustrates the inherent visual quality differences that has also been quantified in our user study (see Fig.~\ref{fig:userstudy}).
The figure also includes the quantitative synchronicity measurements using SyncNet.
Especially, Vougioukas et al.~\cite{Vougioukas2019RealisticSF} achieves a high confidence score, while our method is in the range of the target video it has been trained on (compare to Fig.~\ref{fig:userstudy}). \\

\begin{figure}
    \centering
    \includegraphics[width=0.85\linewidth]{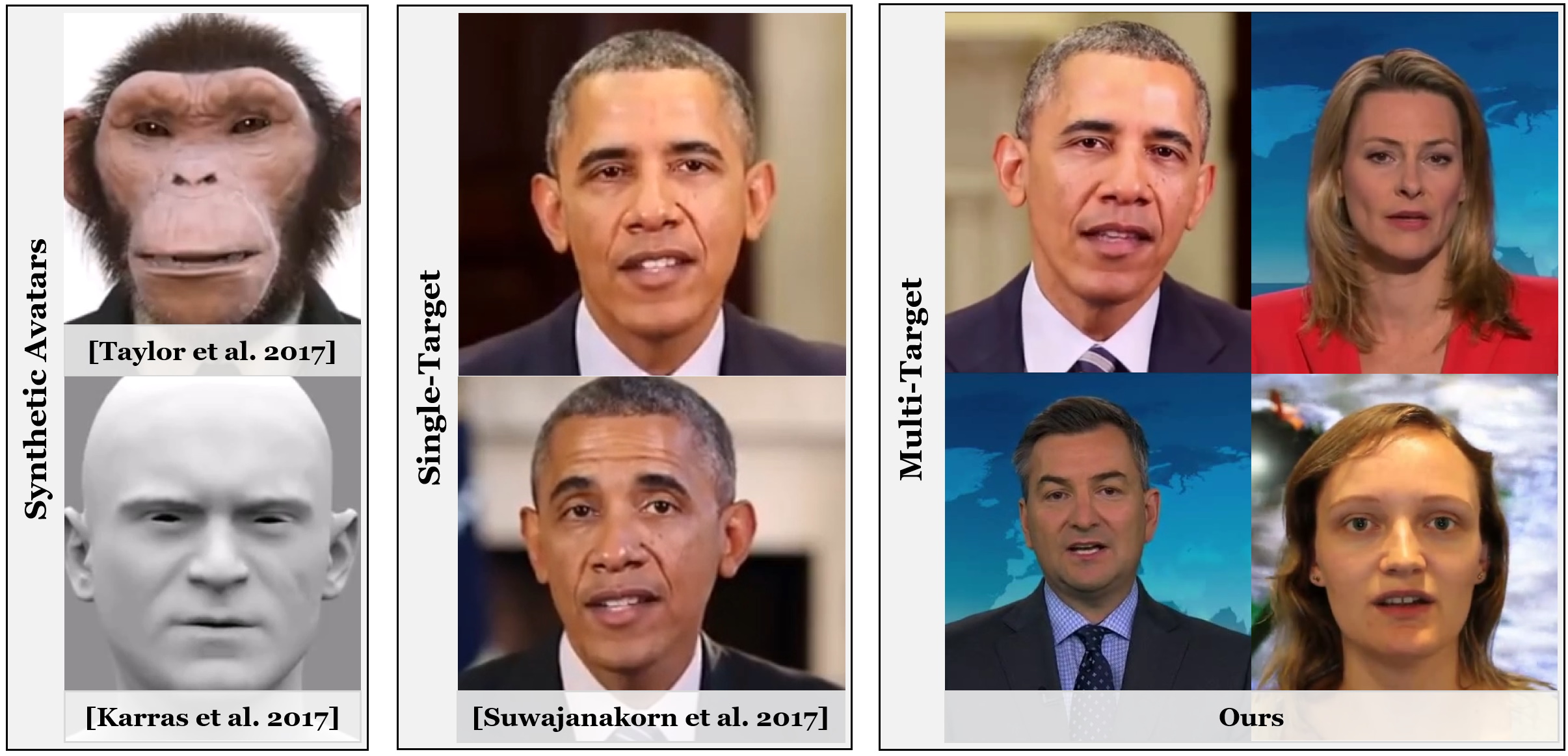}
    \caption{Comparison to state-of-the-art audio-driven model-based video avatars using the same input audio stream. Our approach is applicable to multiple targets, especially, where only $2-3$min of training data are available.}
    \label{fig:comp_model_based_siggraph}
\end{figure}

\noindent
\textbf{Model-based Audio-Driven Methods:}
In Fig.~\ref{fig:comp_model_based_siggraph} we show a representative image from a comparison to Taylor et al.~\cite{Taylor2017}, Karras et al.~\cite{Karras2017} and Suwajanakorn et al.~\cite{Suwajanakorn2017}.
Only the method of Suwajanakorn et al. is able to produce photo-realistic output.
The method is fitted to the scenario where a large video dataset of the target person is available and, thus, limited in its applicability.
They demonstrate it on sequences of Obama, using $14$ hours of training data and $3$ hours for validation. 
In contrast, our method works on short $2-3$ min target video clips.
Measuring the audio visual sync with SyncNet~\cite{syncnet}, the generated Obama videos in Fig.~\ref{fig:comp_model_based_siggraph} (top row) result in $-2$ / $5.9$ (offset/confidence) for the person-specific approach of Suwajanakorn et al., and $0$ / $5.2$ for our generalized expression prediction network.
%


\section{Limitations}
As can be seen in the supplemental video, our approach works robustly on different audio sources and target videos.
But it still has limitations.
Especially, in the scenario of multiple voices in the audio stream our method fails.
Recent work is solving this 'cocktail party' issue using visual cues~\cite{Ephrat2018}.
As all other reenactment approaches, the target videos have to be occlusion free to allow good visual tracking.
In addition, the audio-visual sync of the original videos has to be good since it transfers to the quality of the reenactment.
We assume that the target actor has a constant talking style during a target sequence.
In follow-up work, we plan to estimate the talking style from the audio signal to adaptively control the expressiveness of the facial motions.

\section{Conclusion}
\label{sec:discussion}

We presented a novel audio-driven facial reenactment approach that is generalized among different audio sources.
This allows us not only to synthesize videos of a talking head from an audio sequence from another person, but also to generate a photo-realistic video based on a synthesized voice.
I.e., text-driven video synthesis can be achieved that is in sync with artificial voice.
We hope that our work is a stepping stone in the direction to photo-realistic audio-visual assistants.

\section*{Acknowledgments}
We gratefully acknowledge the support by the AI Foundation, Google, Sony, a TUM-IAS Rudolf M\"o{\ss}bauer Fellowship, the ERC Starting Grant \textit{Scan2CAD} (804724), the ERC Consolidator Grant \textit{4DRepLy} (770784), and a Google Faculty Award. 
%

%
%
\bibliographystyle{splncs04}
\bibliography{paper}

\begin{appendix}

\section{Network Architectures}
\label{sec:net_architecutre}

\subsubsection{Audio2ExpressionNet:}
A core component of \OURS{} is the estimation of facial expressions based on audio.
To retrieve temporal coherent estimations, we employed a process with two stages.
In the first stage, we estimate per frame expressions based on DeepSpeech features.
The output of this network is an audio-expression vector of length $32$.
This audio-expression is temporally noisy and is filtered using an expression aware filtering network which can be trained in conjunction with the per frame expression estimation network.
The temporal filtering mechanism is also depicted in the main paper.
%

\begin{figure}
    \centering
    \includegraphics[width=0.2\textwidth]{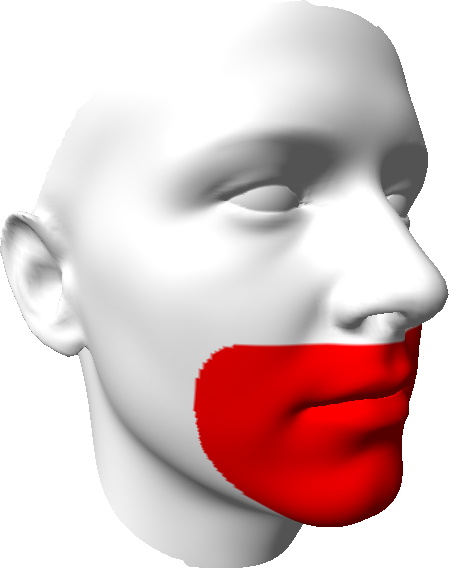}
    \caption{Mask used for the training of the Audio2ExpressionNet.}
    \label{fig:mask}
\end{figure}

%
The underlying network that predicts the filter weights gets as input $T=8$ per-frame predicted audio expressions.
We apply $5$ 1D-convolutional filters with kernel size $3$ that reduce the feature space successively from $8\times32$ over $8\times16$, $8\times8$, $8\times4$, $8\times2$ to $8\times1$.
Each of these convolutions has a bias and is followed by a leaky ReLU activation (negative slope of $0.02$).
The output of the convolutional network is input to a fully connected layer with bias that maps the $1\times8$ input to the $8$ filter weights that are normalized using a softmax function.
To train the network we apply a vertex-based loss as described in the main paper.
The vertices that refer to the mouth region are weighted with a $10\times$ higher loss.
We use the mask that is depicted in Fig.~\ref{fig:mask}.
For generalization we used a dataset composed of commentators from the German public TV (e.g., \url{https://www.tagesschau.de/multimedia/video/video-587039.html}).
In total the dataset contained $116$ videos.

\subsubsection{Rendering network:}
\begin{figure}[h]
    \centering
    \includegraphics[width=\linewidth]{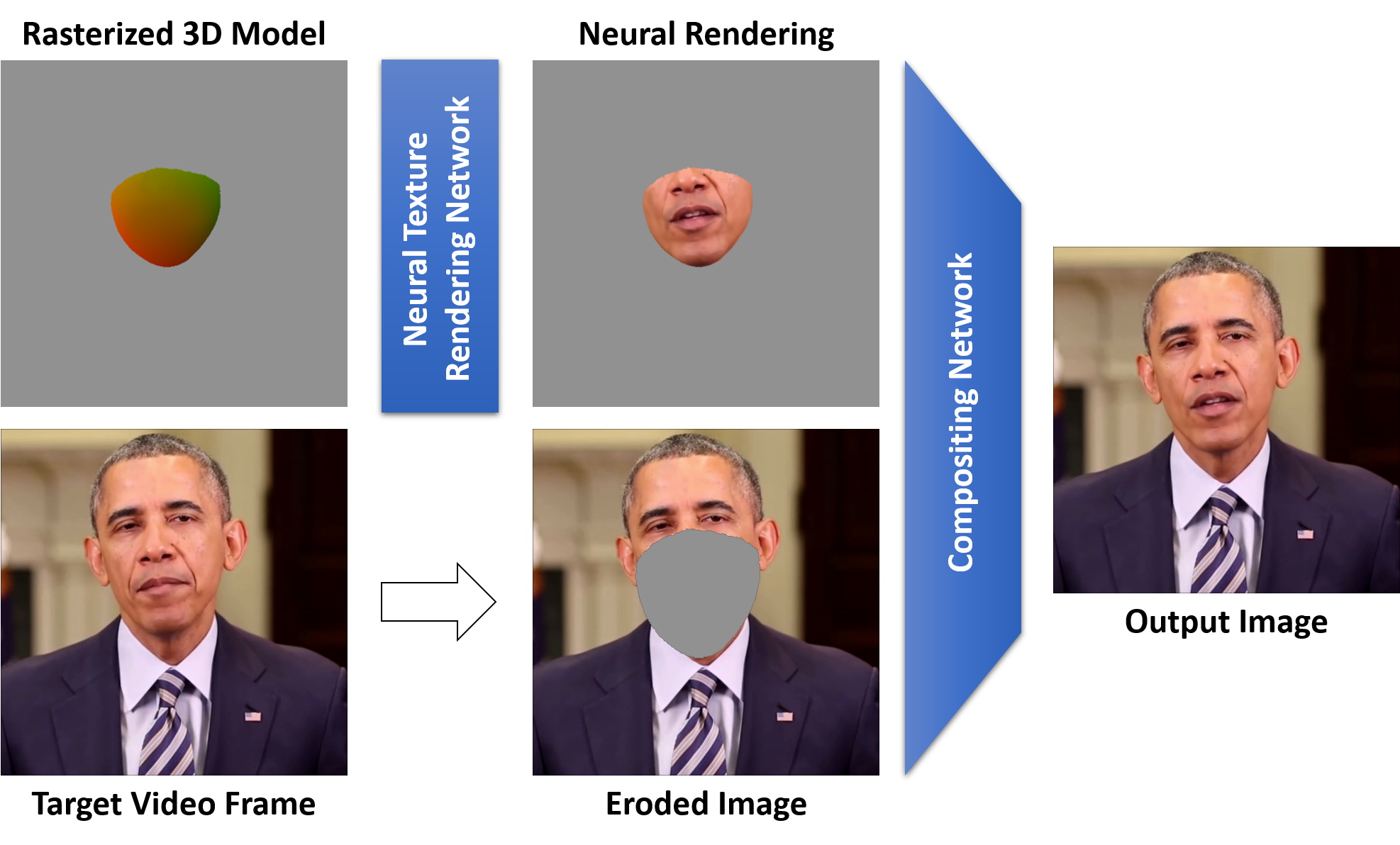}
    \caption{Our neural rendering approach consists of a deferred neural renderer and an inpainting network that blends the modified face interior into the target image.}
    \label{fig:neural_rendering_net}
\end{figure}

In Fig.~\ref{fig:neural_rendering_net}, we show an overview of our neural rendering approach.
Based on the expression predictions, that drive a person-specific 3D face model, we render a neural texture to the image space of the target video.
A first network is used to convert the neural descriptors sampled from the neural texture to RGB color values.
A second network embeds this image into the target video frame.
We erode the target image around the synthetic image, to remove motions of the target actor like chin movements.
Using this eroded target image as background and the output of the first network, the second network outputs the final image.

\begin{figure}
\centering
\includegraphics[width=0.55\textwidth]{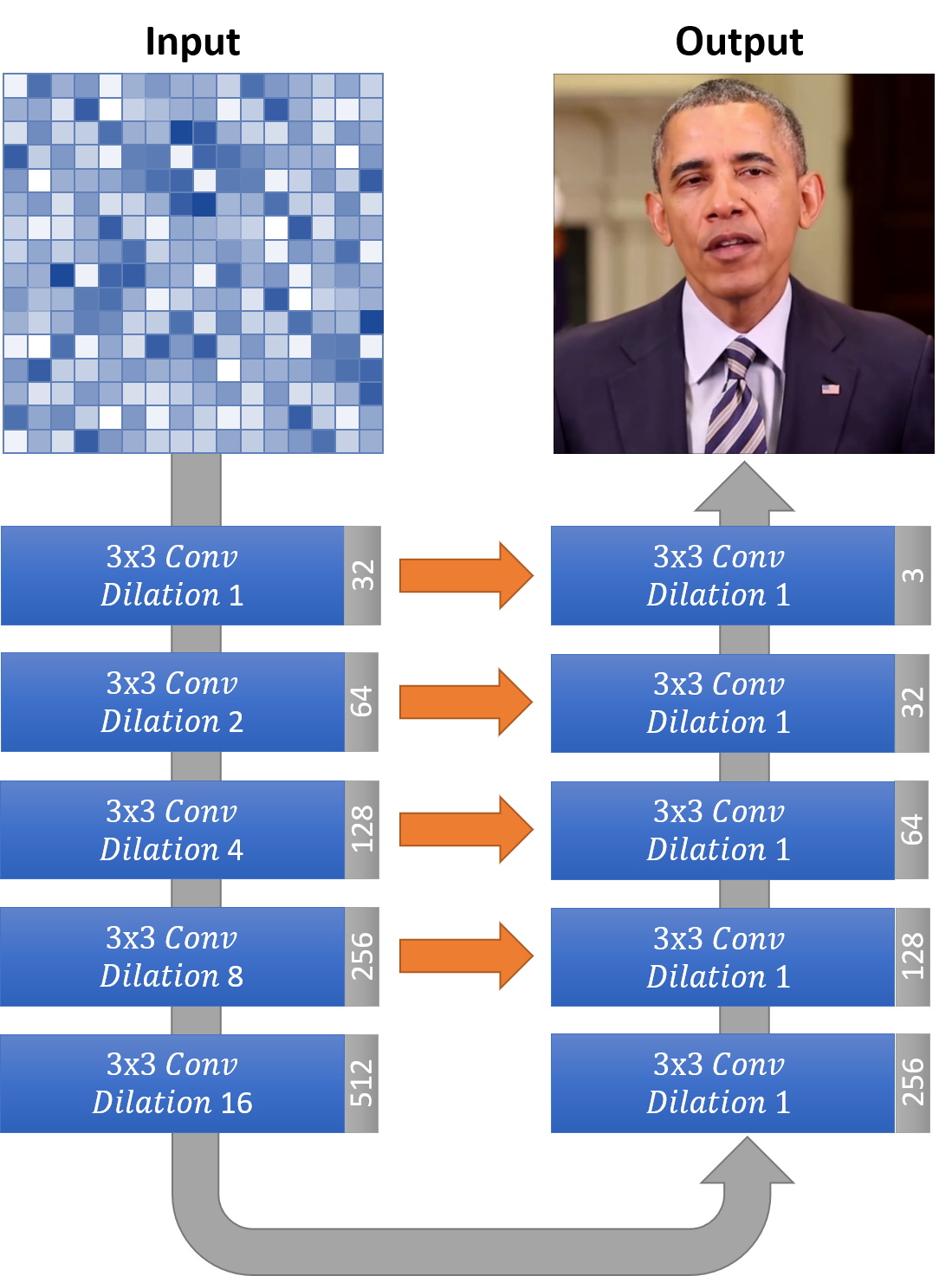}
\caption{We use a modified U-Net architecture that uses dilated convolutions instead of strided convolutions. Transposed convolutions are replaced by std. convolutions.}
\label{fig:renderer_net}
\end{figure}

Both networks have the same structure, only the input dimensions are different.
The first network gets an image with $16$ feature channels as input (dimension of the neural descriptors that are sampled from a neural texture with dimensions $256\times256\times16$), while the second network composites the background and the output of the first network, resulting in an $6$ channel input.
The networks are implemented in the Pix2Pix framework~\cite{pix2pix}.
Instead of a classical U-Net with strided convolutions, we build on dilated convolutions.
Specifically, we replace the strided convolutions in a U-Net of depth $5$.
Instead of transposed convolutions, we use standard convolutions, since we do not downsample the image and always keep the same image dimensions.
Note that we also keep the skip connections of the classical U-Net.
The number of features per layer is $32$ in our experiments, resulting in networks with $\sim2.35mio$ parameters (which is low in comparison to the network in Deferred Neural Rendering~\cite{thies2019} with $\sim16mio$ parameters).
We employ the structure that is depicted in Fig.~\ref{fig:renderer_net}.
Each convolution layer has a kernel size of $3\times3$ and is followed by a leaky ReLU with negative slop of $0.2$.
All layers have stride $1$ which means that all layers intermediate feature maps have the same spatial size as the input ($512\times512$).
The first convolutional layer maps to a feature space of dimension $32$ and has a dilation of $1$.
With increasing layer depth the feature space dimension as well as the dilation increases by a factor of $2$.
After layer depth $5$, we use standard convolutions.

\subsubsection{Training}
Our pipeline is implemented in PyTorch using the Adam~\cite{adam} optimizer with default settings ($\beta_1=0.9$, $\beta_2=0.999$, $\epsilon=1\cdot e^{-8}$), a learning rate of $0.0001$ and Xavier initialization.
The \textit{Audio2ExpressionNet} is trained for $50$ epochs (resulting in a training time of $\sim28$ hours on an Nvidia 1080Ti) with a learning rate decay for the last $30$ epochs and a batch size of $16$.
The rendering networks are trained for $50$ epochs for each target person individually with a batch size of $1$ ($\sim30$ hours training time, $\sim5$ hours in case of strided convolutions).

\section{User Study}
\label{sec:userstudy}

\begin{figure*}
    \centering
    \includegraphics[width=\linewidth]{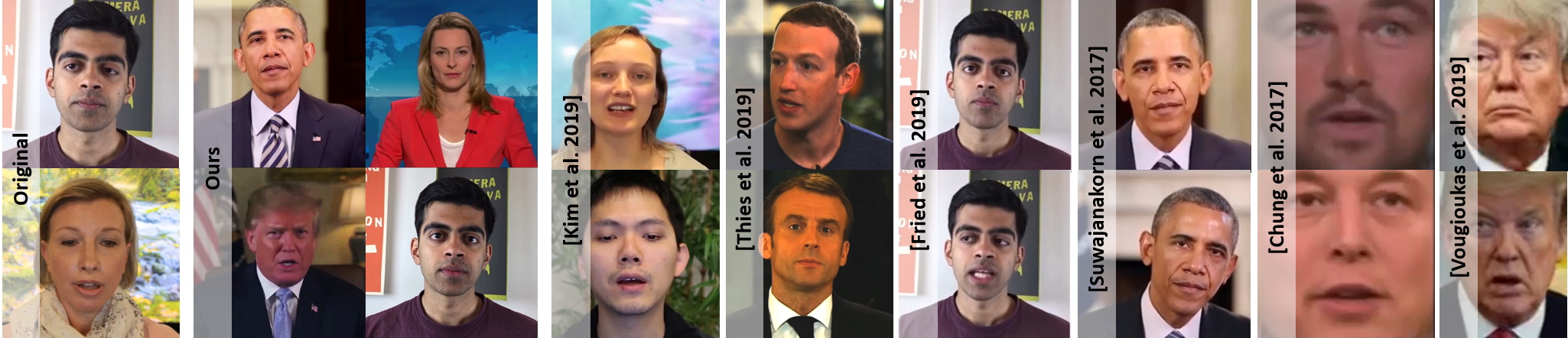}
    \caption{Our user study contained $24$ videos from different state-of-the-art methods, including $3$ original videos. Here we show some frames of the videos.}
    \label{fig:userstudy_videos}
\end{figure*}

In this section, we present the statistics of our user study.
Fig.~\ref{fig:userstudy_videos} shows a collection of videos that we used for the user study.
The clips are from the official videos of the corresponding methods and are similar to the clips that we show in our supplemental video.
Fig.~\ref{fig:userstudy_videos} shows the average answers of our questions, including the variance.
\\ \\
\noindent
In the user study we asked the following questions:
\begin{itemize}
    \item How would you rate the audio-visual alignment (lip sync) in this video?
    \item How would you rate the visual quality of the video?
\end{itemize}
With the answer possibilities "very good","good","Neither good nor bad","bad", "very bad".

\begin{figure*}
    \centering
    \includegraphics[width=0.75\linewidth]{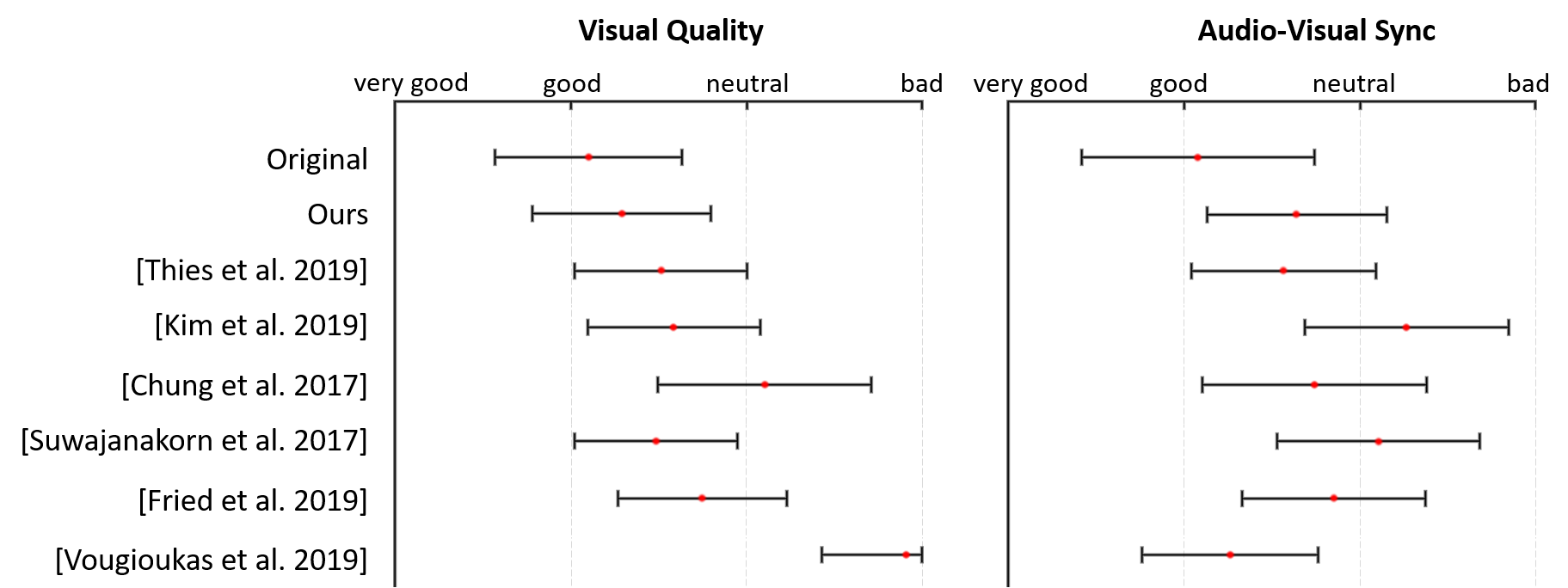}
    \caption{Statistics of our user study including the mean and the variance with respect to the specific methods and question about visual quality and audio-visual sync (lip sync).}
    \label{fig:userstudy_videos}
\end{figure*}

\section{Additional Comparisons to State-of-the-Art}

\subsubsection{Image-based \& Audio-driven Facial Animation:}
\begin{figure*}
    \centering
    \includegraphics[width=1.0\linewidth]{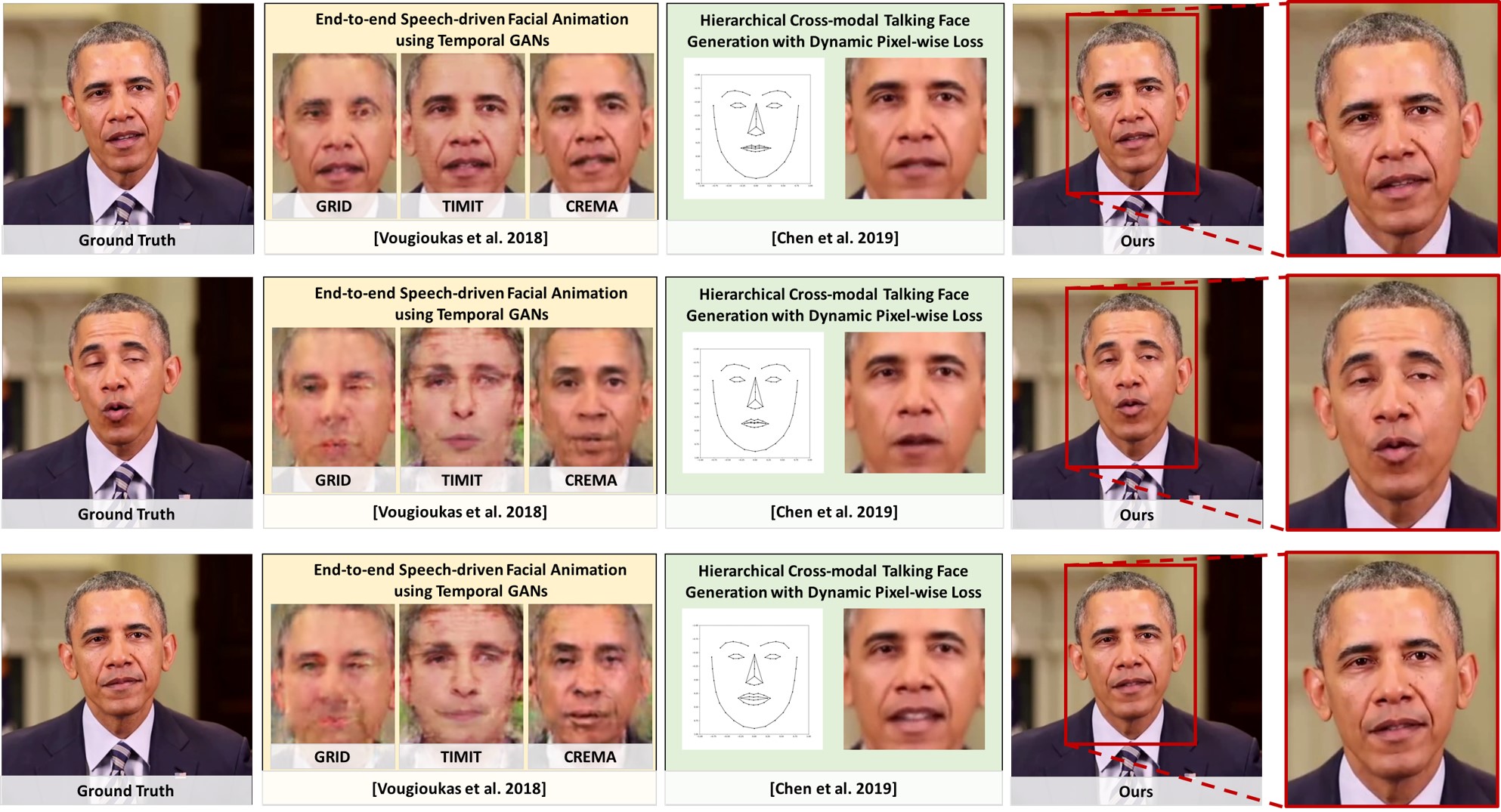}
    \caption{Comparison to Chen et al.~\cite{chen2019hierarchical} and Vougioukas et al.~\cite{Vougioukas2018EndtoEndSF} on a sequence of Obama (self-reenactment). From top to bottom we show the frames $0$, $70$ and $305$. Note that we list $3$ results for the method of Vougioukas et al. which are based on different training datasets (GRID, TIMIT, CREMA). The respective sequence is part of the supplemental video. }
    \label{fig:additional_comp}
\end{figure*}

In addition to the results in the main paper, we also compare to Chen et al.~\cite{chen2019hierarchical} and Vougioukas et al.~\cite{Vougioukas2018EndtoEndSF}.
For both methods, we use publicly available pretrained models~\footnote{\url{https://github.com/lelechen63/ATVGnet}}\footnote{\url{https://github.com/DinoMan/speech-driven-animation}}.
We compare on a sequence of Obama in a self-reenactment scenario (to provide ground truth images).
As can be seen in Fig.~\ref{fig:additional_comp}, our method surpasses the visual image quality of these methods and generates full frame images (in contrast to normalized facial images).
Since the image-based methods are operating in a normalized space, we cannot provide a fair quantitative evaluation w.r.t. the ground truth images.
To compute PSNR and landmark errors, we would need to transform the ground truth images to the normalized space which leads to errors since the head is moving and we can not assume perfect tracking of the face bounding box.
Especially, rotations of the head are also not handled by the image-based methods which would dominate the error.
Our PSNR and landmark errors for the self-reenactment sequence of Obama are listed in the main paper.

\subsubsection{Model-based Audio-driven Facial Animation:}
\begin{figure}
    \centering
    \includegraphics[width=0.7\linewidth]{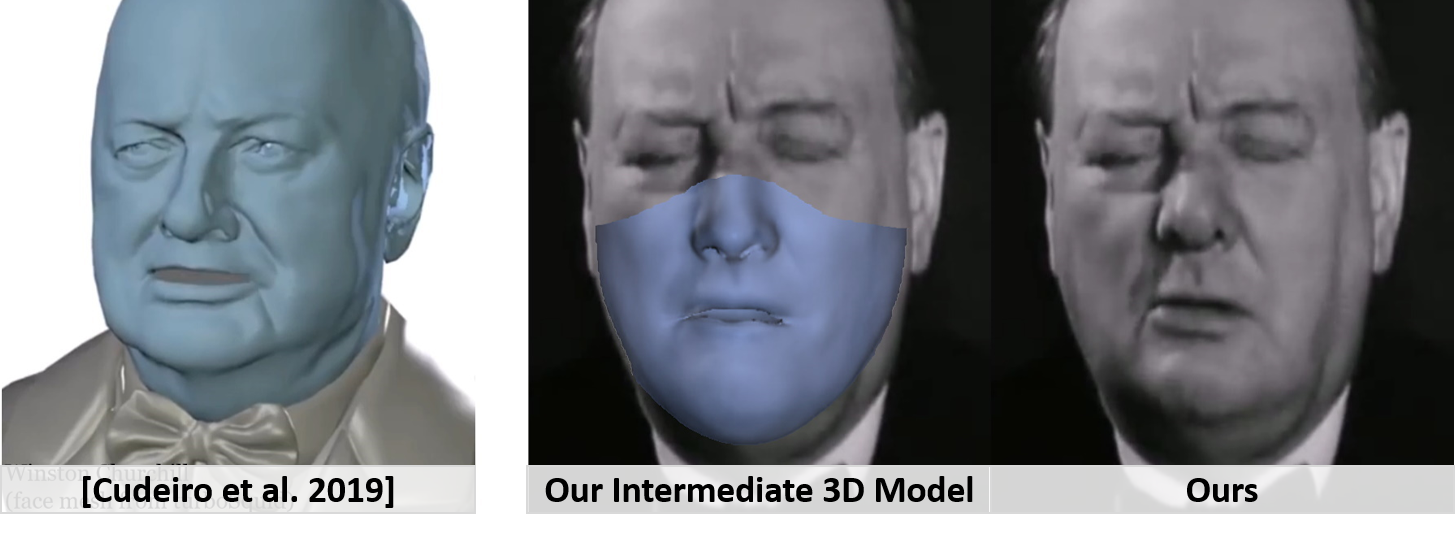}
    \caption{Qualitative comparison of our method to Voca~\cite{VOCA2019}. It is a representative image for a talking sequence of Winston Churchill.}
    \label{fig:comp_voca}
\end{figure}

In our supplemental video, we show multiple comparisons to Voca~\cite{VOCA2019}.
Fig.~\ref{fig:comp_voca} shows an image of a legacy Winston Churchill sequence.
In contrast to Voca, our aim is to generate photo-realistic output videos that are in sync with the audio. 
Voca focuses on the 3D geometry requiring a 4D training corpus, while our approach uses a 3D proxy only as an intermediate step and works on videos from the Internet.
Our 3D proxy is based on a generic face model and, thus, has not the details as a person-specific modelled mesh.
Nevertheless, using our neural rendering approach, we are able to generate photo-realistic results.

\subsubsection{Model-based Video-driven Dubbing \& Facial Reenactment:}
\begin{figure}
    \centering
    \includegraphics[width=0.7\linewidth]{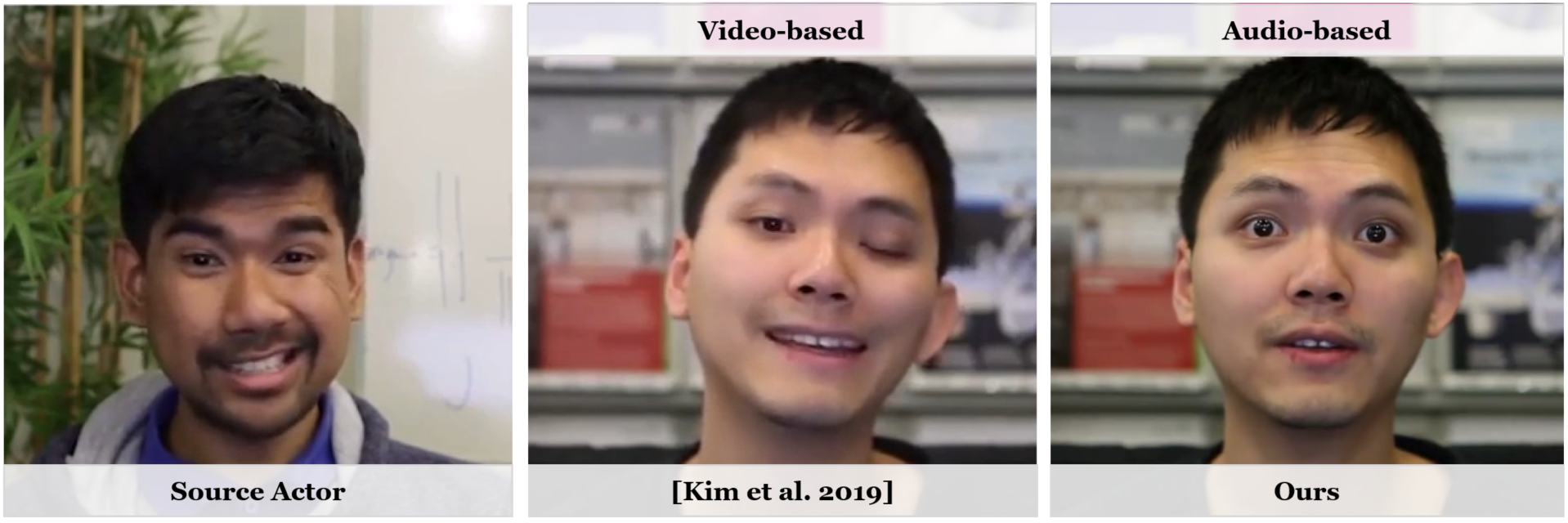}
    \caption{Visual dubbing fails to map strong expressions from the source to plausible expressions of the target actor.}
    \label{fig:neural_dubbing}
\end{figure}

State-of-the-art video dubbing is based on video-driven facial reenactment~\cite{Garrido2015,thies2016face,kim2018DeepVideo,thies2019,Kim19NeuralDubbing}.
In contrast, our method is only relying on the voice of the dubber.
The 'Deferred Neural Rendering'~\cite{thies2019} is a generic neural rendering approach, but the authors also show the usage in the scenario of facial reenactment.
It builds upon the Face2Face~\cite{thies2016face} pipeline and directly transfers the deformations from the source to the target actor.
Thus, tracking errors that occur in the source video (e.g., due to occlusions or fast motions) are transferred to the target video.
In a dubbing scenario, the goal is to keep the talking style of the target actor which is not the case for \cite{Garrido2015,thies2016face,kim2018DeepVideo,thies2019}.
To compensate the influence of the source actor talking style, Kim et al.~\cite{Kim19NeuralDubbing} proposed a method to map from the source style to the target actor style.
Our approach directly operates in the target actor expression space, thus, no mapping is needed (we also do not capture the source actor style).
This enables us to also work on strong expressions, as shown in Fig.~\ref{fig:neural_dubbing}.

\subsubsection{Text-driven Video Synthesis:}
Fried et al. presented 'Text-based Editing of Talking-head Video'~\cite{Fried2019} which provides a video editing tool that is based on the transcript of the video.
The method reassembles captured expression snippets from the target video, requiring blending heuristics.
To achieve their results they rely on more than one hour of training data.
We show a qualitative comparison to this method in the supplemental video.
Our method only uses the synthetic audio sequence as input, while the method of Fried et al. uses both the transcript and the audio.
Note that our method generates the entire video, while the text-based editing method only synthesizes the frames of the new three words.

\section{Ethical Considerations}

In conjunction with person specific audio generators like Jia et al.~\cite{Jia2018}, a pipeline can be established that creates video-realistic (temporal voice- and photo-realistic) content of a person.
This is perfect for creative people in movie and content production, to edit and create new videos.
On the other hand, it can be misused.
To this end, the field of digital media forensics is getting more attention.
Recent publications~\cite{roessler2019faceforensics++} show that humans have a hard time in detecting fakes, especially, in the case of compressed video content.
Learned detectors are showing promising results, but are lacking generalizeability to other manipulation methods that are not in the training corpus.
Few-shot learning methods like ForensicTransfer~\cite{cozzolino2018forensictransfer} try to solve this issue.
As part of our responsibility, we are happy to share generated videos of our method with the forensics community.
Nevertheless, our approach enables several practical use-cases, ranging from movie-dubbing to text-driven photo-realistic video avatars.
We hope that our work is a stepping stone in the direction of audio-based reenactment and is inspiring more follow-up projects in this field. 

\end{appendix}

\end{document}